\documentclass[12pt]{article}

\usepackage{makecell}
\usepackage{amssymb,amsmath,amsfonts,latexsym,graphicx}
\usepackage[numbers]{natbib}
\usepackage{multirow}
\usepackage{hyperref} 
\hypersetup{
    colorlinks=true,
    urlcolor=blue,
    citecolor=black,
    pdftitle={Overleaf Example},
    pdfpagemode=FullScreen,
}

\begin{document}

\begin{center}
\Large \bf VISTA: Validation-Guided Integration of Spatial and Temporal Foundation Models with Anatomical Decoding for Rare-Pathology VCE Event Detection \textcolor{blue}{- after competition results} \rm







\vspace{1cm}


\large Bo-Cheng Qiu $\,^a$, \large Fang-Ying Lin $\,^b$, \large Ming-Han Sun $\,^a$, \large Yu-Fan Lin $\,^a$, \large Chia-Ming Lee $\,^a$, \large Chih-Chung Hsu $\,^b,^*$


\vspace{0.5cm}

\normalsize


$^a$ National Cheng Kung University, Taiwan

$^b$ National Yang Ming Chiao Tung University, Taiwan

\vspace{5mm}


Corresponding Author Email: {\tt \href{mailto:chihchung@nycu.edu.tw}{chihchung@nycu.edu.tw}} 

Team Name: {\tt ACVLab}

GitHub Repository Link: {\tt \href{https://github.com/Joe1007/ICPR-2026-RARE-VISION-Challenge}{https://github.com/Joe1007/ICPR-2026-RARE-VISION-Challenge}}

\vspace{1cm}

\end{center}

\abstract{Capsule endoscopy event detection is challenging because clinically relevant findings are sparse, visually heterogeneous, and evaluated at the event level rather than by frame accuracy. We propose VISTA, a metric-aligned multi-backbone framework for the RARE-VISION task. VISTA combines EndoFM-LV for temporal context and DINOv3 ViT-L/16 for frame-level visual semantics, followed by a Diverse Head Ensemble (DHE), Validation-Guided Weighted Fusion (VGWF), and Anatomy-Aware Temporal Event Decoding (ATED). The original official submission achieved hidden-test temporal mAP@0.5 of 0.3530 and mAP@0.95 of 0.3235. After the competition, extending local threshold refinement with a global coarse search improved performance to 0.3726 mAP@0.5 and 0.3431 mAP@0.95, ranking Team ACVLab second in the post-competition evaluation.}

\section{Motivation}\label{sec1}

Capsule endoscopy analysis is difficult because clinically relevant findings are sparse, visually heterogeneous, and distributed across long video recordings. The Galar dataset and the RARE-VISION challenge reflect this setting through large-scale multi-label video capsule endoscopy data, severe class imbalance, rare pathological findings, and event-level evaluation based on temporal mAP rather than frame accuracy alone \cite{LeFloch2025figshareplus,Lawniczak2025}. Related gastrointestinal image grounding work further shows that clinically relevant findings require robust recognition and localization under complex visual conditions \cite{lin2024divide}. Consequently, strong frame-wise predictions may still lead to weak final performance if they are temporally fragmented, poorly thresholded, or anatomically inconsistent.

These challenges require both temporal context and strong frame-level visual semantics. Long-sequence representation learning is useful for endoscopy videos because temporal context improves downstream video understanding \cite{wang2025improving}. Meanwhile, robust medical visual recognition requires discriminative features that are less sensitive to source variation \cite{lee2025taming}, and large self-supervised vision models provide strong dense visual representations without task-specific backbone redesign \cite{simeoni2025dinov3}. This motivates our dual-backbone design, which combines an endoscopy video foundation model for temporal dynamics with a general visual foundation model for anatomical and pathological frame-level semantics.

The long-tailed multi-label setting further requires imbalance-aware learning and label-sensitive aggregation. Asymmetric loss improves robustness for rare tail classes in multi-label long-tailed settings \cite{park2023robust}, while weighted strategies based on class frequency and disease co-occurrence improve rare disease discrimination in multi-label medical imaging \cite{lin2025weighted}. We therefore use heterogeneous loss functions across ensemble heads and validation-guided class-wise weighting during fusion.

Finally, robust inference is critical because temporal mAP evaluates event segments rather than isolated frame labels. Confidence adjustment, threshold selection, and temporal refinement are therefore needed to convert noisy frame-wise probabilities into stable event predictions. Prior wireless capsule endoscopy work also suggests that temporal information remains useful when visual evidence is degraded \cite{nam2024deep}. Based on these considerations, we formulate the task as a metric-aligned event detection pipeline that combines complementary backbones, diverse ensemble heads, validation-guided weighted fusion, and anatomy-aware temporal decoding.

\section{Methods}\label{sec2}

\begin{figure}[htbp]
    \centering
    \includegraphics[width=0.95\linewidth]{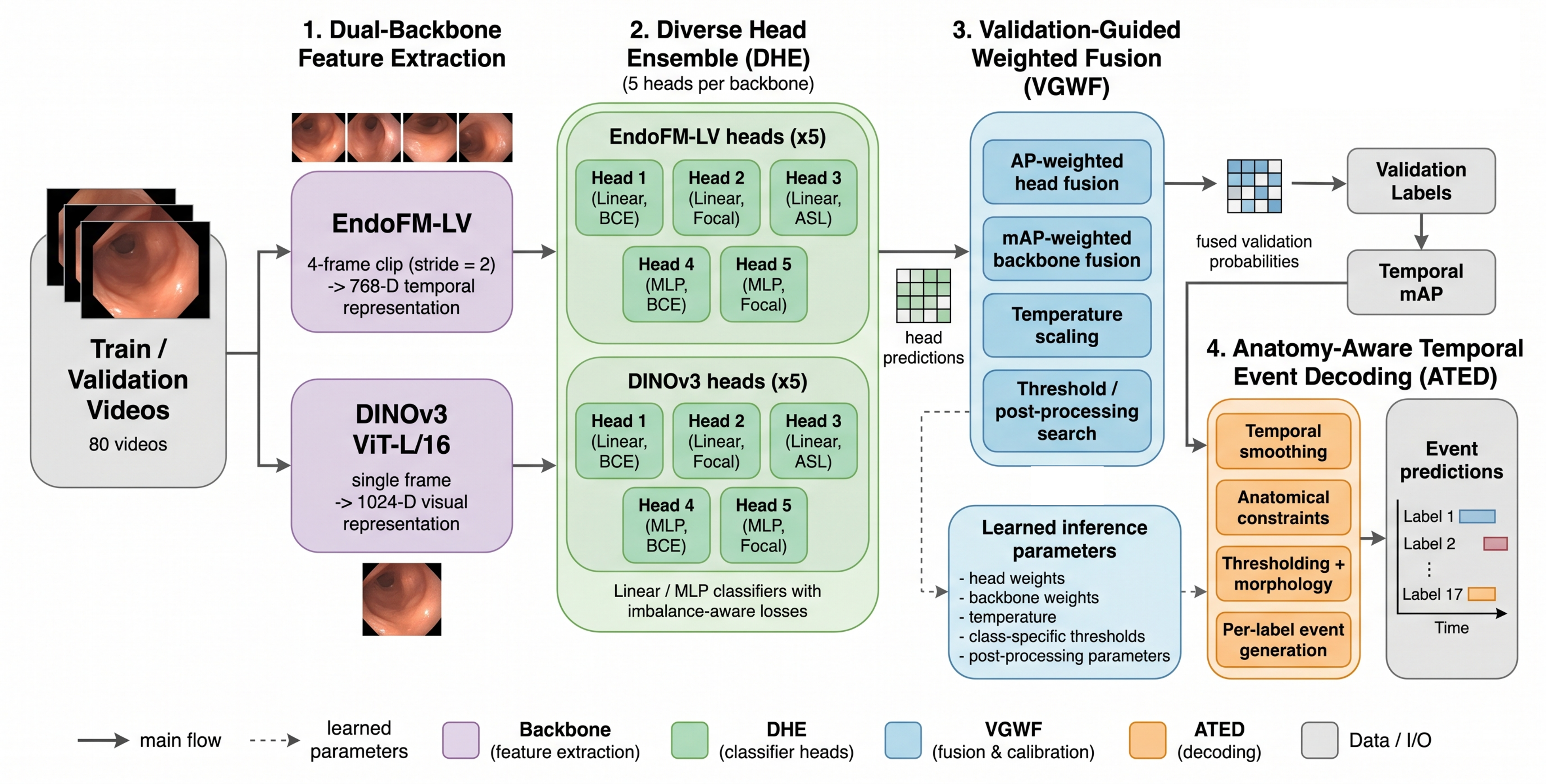}
    \caption{Overview of the developed pipeline.}
    \label{fig:enter-label}
\end{figure}

Figure~\ref{fig:enter-label} shows the overall pipeline. We extract complementary temporal and visual features using two backbones, train a Diverse Head Ensemble (DHE) on each feature stream, fuse predictions by Validation-Guided Weighted Fusion (VGWF), and convert fused frame-level probabilities into event predictions through Anatomy-Aware Temporal Event Decoding (ATED).

\subsection{Dual-Backbone Feature Extraction}
We use two complementary backbones on the same train/validation split. EndoFM-LV encodes short-range temporal context from a 4-frame clip with stride 2:
\begin{equation}
h_t^{(1)} = f_{\text{EndoFM}}(c_t),
\end{equation}
where $h_t^{(1)} \in \mathbb{R}^{768}$. DINOv3 ViT-L/16 extracts a frame-level visual representation
\begin{equation}
h_t^{(2)} = f_{\text{DINOv3}}(x_t),
\end{equation}
where $h_t^{(2)} \in \mathbb{R}^{1024}$. The two backbones provide complementary temporal and frame-level visual cues.

\subsection{Diverse Head Ensemble and Validation-Guided Weighted Fusion}
On top of each backbone, we train a \textbf{Diverse Head Ensemble (DHE)} with five lightweight classifiers using different architectures and loss functions, including linear or MLP heads optimized by BCE, focal loss, or asymmetric loss. For backbone $b$, head $m$, and class $c$, the predicted probability is denoted by $p_{t,b,m,c}$.

These outputs are fused by \textbf{Validation-Guided Weighted Fusion (VGWF)}. For each backbone, class-wise Average Precision (AP) on the validation set determines classifier-head weights. Let $b \in \{1,\ldots,B\}$ denote the backbone index, $m \in \{1,\ldots,M_b\}$
denote the classifier-head index for backbone $b$, and $c \in \{1,\ldots,C\}$
denote the class index.
\begin{equation}
\alpha_{b,m,c}
=
\frac{\mathrm{AP}_{b,m,c}}
{\sum_{m'=1}^{M_b} \mathrm{AP}_{b,m',c}},
\qquad
\hat{p}_{t,b,c}
=
\sum_{m=1}^{M_b} \alpha_{b,m,c} p_{t,b,m,c}.
\end{equation}

The two backbone-level predictions are then fused using validation frame-level mAP:
\begin{equation}
\beta_b
=
\frac{\mathrm{mAP}_b}
{\sum_{b'=1}^{B}\mathrm{mAP}_{b'}},
\qquad
\hat{p}_{t,c}
=
\sum_{b=1}^{B}\beta_b \hat{p}_{t,b,c}.
\end{equation}
We further adjust the fused probabilities by temperature scaling,
\begin{equation}
\tilde{p}_{t,c}=\sigma\!\left(\frac{\mathrm{logit}(\hat{p}_{t,c})}{T}\right),
\end{equation}
where $T$ is selected by grid search on the validation set.

The same validation stage is also used to tune downstream inference parameters. Class-specific thresholds are initialized from precision--recall curves using F1 and refined by local search with temporal mAP as the target objective. Temporal smoothing and morphological refinement parameters are also selected on the validation set.

\subsection{Anatomy-Aware Temporal Event Decoding}
At test time, EndoFM-LV extracts clip-level features online, with predictions averaged
over original and horizontally flipped clips as a simple test-time augmentation, while
DINOv3 uses pre-extracted frame features. Predictions are fused using the validation-derived classifier-head weights, backbone weights, temperature, thresholds, and post-processing parameters.

The fused probability sequence is then processed by \textbf{Anatomy-Aware Temporal Event Decoding (ATED)}. First, class-specific temporal smoothing is applied independently to each video, with larger windows for anatomical classes and smaller windows for pathological classes. Second, we apply simple anatomical constraints during decoding. The five major anatomical regions are treated as mutually exclusive, so only the highest-probability region is retained at each frame. We also enforce the monotonic gastrointestinal transit order from \texttt{mouth} to \texttt{colon}, which suppresses anatomically implausible backward transitions. In addition, landmark labels are only retained when they appear within, or close to, their anatomically valid neighboring regions. Third, class-specific thresholds are applied to obtain binary sequences, followed by morphological opening and closing to suppress short false positives and reconnect fragmented events. If all anatomical regions are removed at a frame, the most likely region is reassigned to ensure full anatomical coverage.

Finally, event generation is performed independently for each label, where each contiguous positive segment is treated as one event. This per-label decoding is more faithful to temporal IoU-based evaluation than tuple-based multi-label segmentation because pathological events are not artificially split when anatomical labels change over time.

\subsection{How was class imbalance handled?}
Class imbalance is handled at both training and inference time. During training, we use BCE with \texttt{pos\_weight}, focal loss, and asymmetric loss across different ensemble heads to improve learning for rare labels. During inference, imbalance is further mitigated by class-wise ensemble weighting and class-specific threshold optimization, allowing minority pathological classes to benefit from models and operating points with better recall.

\subsection{\textcolor{blue}{Post competition changes in the Methods}}
After the competition, we refined the validation-guided threshold search while keeping the trained backbones, ensemble heads, fusion strategy, and ATED pipeline unchanged. In the original implementation, each class-specific threshold was initialized from the precision--recall curve using F1 and then refined only within a local range around the F1-derived threshold $b_c$, i.e., $[b_c-0.15, b_c+0.15]$. This can miss better event-level operating points when $b_c$ is close to one end of the probability range.

To reduce this dependence on the initial threshold, we added a coarse global search over $[0.05,0.95]$ and selected the threshold that maximized validation temporal mAP@0.5 after the same post-processing steps. The final candidate set for class $c$ is

\[
\mathcal{T}^{\mathrm{thr}}_c =
\mathrm{Local}(b_c, 0.15) \cup \mathrm{Global}(0.05, 0.95),
\tag{6}
\]

where $b_c$ denotes the F1-derived initial threshold. This modification does not use additional training data or change learned model parameters. It improved the hidden-test performance from 0.3530 to 0.3726 in temporal mAP@0.5 and from 0.3235 to 0.3431 in temporal mAP@0.95, mainly due to improved performance on \texttt{ukdd\_navi\_00051}.

\section{Results}\label{sec3}

\paragraph{Dataset and split.}
We follow the official ICPR 2026 RARE-VISION competition protocol based on the Galar dataset. The development cohort contains 80 labeled videos with frame-wise annotations and metadata, and is used for model training and validation. Following the competition setting, the official test set is strictly separated from the development data and consists of 3 previously unseen NaviCam examinations with withheld annotations for blind evaluation. As the competition protocol does not prescribe a fixed internal train/validation partition within the 80 development videos, validation was performed using our own split strategy on the released development cohort.

We report validation-set ablations using the official event-level metrics temporal mAP@0.5 and temporal mAP@0.95. Table~\ref{tab:ablation_main} evaluates the main design choices of the proposed framework, including complementary backbones, validation-guided weighted fusion (VGWF), and anatomy-aware temporal event decoding (ATED). The full method achieved the best validation result among all evaluated configurations, and we can conclude three main observations.

\begin{table*}[htbp]
\centering
\small
\setlength{\tabcolsep}{1pt}
\renewcommand{\arraystretch}{1.15}
\begin{tabular}{l l l c c}
\hline
\textbf{Backbone(s)} & \textbf{Fusion} & \textbf{Event decoding} & \textbf{mAP@0.5} & \textbf{mAP@0.95} \\
\hline

EndoFM-LV 
& Single backbone 
& \makecell[l]{Full ATED \\ + per-label events} 
& 0.4504 & 0.3873 \\

DINOv3 
& Single backbone 
& \makecell[l]{Full ATED \\ + per-label events} 
& 0.5055 & 0.4091 \\

\hline

EndoFM + DINOv3 
& \makecell[l]{AP-weighted classifier head \\ + mAP-weighted backbone} 
& Per-label events only 
& 0.4375 & 0.3187 \\

EndoFM + DINOv3 
& \makecell[l]{AP-weighted classifier head \\ + mAP-weighted backbone} 
& \makecell[l]{Full ATED \\ + tuple-based events} 
& 0.4876 & 0.4032 \\

\hline

EndoFM + DINOv3 
& \makecell[l]{Uniform classifier head \\ + Uniform backbone} 
& \makecell[l]{Full ATED \\ + per-label events} 
& 0.5487 & 0.4137 \\

EndoFM + DINOv3 
& \makecell[l]{Uniform classifier head \\ + mAP-weighted backbone} 
& \makecell[l]{Full ATED \\ + per-label events} 
& 0.5382 & 0.4075 \\

EndoFM + DINOv3 
& \makecell[l]{AP-weighted classifier head \\ + Uniform backbone} 
& \makecell[l]{Full ATED \\ + per-label events} 
& 0.5156 & 0.4108 \\

\textbf{EndoFM + DINOv3} 
& \makecell[l]{\textbf{AP-weighted classifier head} \\ \textbf{+ mAP-weighted backbone}} 
& \makecell[l]{\textbf{Full ATED} \\ \textbf{+ per-label events}} 
& \textbf{0.5520} & \textbf{0.4148} \\

\hline
\end{tabular}
\caption{Validation-set ablation study of the proposed framework.}
\label{tab:ablation_main}
\end{table*}

First, the two backbones provide complementary information, although DINOv3 is the stronger single-backbone model. After combining EndoFM-LV and DINOv3 with the full fusion and decoding pipeline, mAP@0.5 further increased to 0.5520 and mAP@0.95 to 0.4148, suggesting that the temporal cues from EndoFM-LV remain useful for event-level detection.

Second, VGWF provides a modest but consistent improvement when applied jointly at both the classifier-head and backbone levels. Uniform classifier-head and uniform backbone fusion already achieved strong performance, with 0.5487 mAP@0.5 and 0.4137 mAP@0.95. Applying AP-weighted classifier-head fusion together with mAP-weighted backbone fusion slightly improved the result to 0.5520 mAP@0.5 and 0.4148 mAP@0.95. In contrast, using only one level of weighting was less effective: uniform classifier-head fusion with mAP-weighted backbone fusion reached 0.5382/0.4075, while AP-weighted classifier-head fusion with uniform backbone fusion reached 0.5156/0.4108. This suggests that VGWF is useful, but its benefit is relatively small and depends on using both weighting levels together.

Third, ATED is important for converting frame-level predictions into event-level outputs. Using AP-weighted classifier-head fusion and mAP-weighted backbone fusion with per-label
event generation but without full ATED achieved 0.4375 mAP@0.5 and 0.3187 mAP@0.95. Adding full ATED with tuple-based event generation improved the result to 0.4876/0.4032, showing that temporal smoothing, anatomical constraints, and morphological refinement substantially improve event-level consistency. Replacing tuple-based event generation with per-label event generation further improved performance to 0.5520/0.4148, indicating that independent per-label event construction better matches the temporal IoU-based evaluation protocol and avoids unnecessary fragmentation when anatomical labels change over time.

\paragraph{Official test-set results.}
We submitted the original full method to the official competition scoring system, and the
hidden-test results are shown as the Before phase in Table~\ref{tab:competition_results}. After the competition, the global-augmented threshold refinement improved the overall
mAP@0.5 from 0.3530 to 0.3726 and mAP@0.95 from 0.3235 to 0.3431.

\begin{table}[htbp]
    \centering
    \caption{\textcolor{blue}{Performance comparison before and after the competition.}}
    \begin{tabular}{c|c|cc}
        \hline
        Video & Phase & mAP@0.5 & mAP@0.95 \\
        \hline
        
        \multirow{2}{*}{ukdd\_navi\_00051}
        & Before & 0.4706 & 0.4412 \\
        & After  & 0.5295 & 0.5000 \\
        \hline
        
        \multirow{2}{*}{ukdd\_navi\_00068}
        & Before & 0.2356 & 0.1765 \\
        & After  & 0.2353 & 0.1765 \\
        \hline
        
        \multirow{2}{*}{ukdd\_navi\_00076}
        & Before & 0.3529 & 0.3529 \\
        & After  & 0.3529 & 0.3529 \\
        \hline
        
        \multirow{2}{*}{Average (3 Videos)}
        & Before & 0.3530 & 0.3235 \\
        & After  & 0.3726 & 0.3431 \\
        \hline
        
        \textbf{\multirow{2}{*}{Overall}}
        & \textbf{Before} & \textbf{0.3530} & \textbf{0.3235} \\
        & \textbf{After}  & \textbf{0.3726} & \textbf{0.3431} \\
        
        \hline
    \end{tabular}
    \label{tab:competition_results}
\end{table}

\section{Discussion}\label{sec4}

Overall, the results suggest that event-level capsule endoscopy detection depends not only on strong visual representations, but also on inference-time conversion from frame-level probabilities to temporally coherent events. The validation ablation shows that DINOv3 provides the strongest single-backbone representation, while combining it with EndoFM-LV improves mAP@0.5 under the full event-decoding pipeline. This indicates that temporal features can complement strong frame-level visual semantics, especially for recovering event-level continuity.

The ablation results also show that the benefit of validation-guided weighted fusion (VGWF) is present but relatively modest. The fully weighted fusion strategy achieved the best validation mAP@0.5 and the best fused-model mAP@0.95, but the difference from uniform classifier-head and uniform backbone fusion was small. This suggests that the major performance gain comes from using strong complementary backbones and structured event decoding, while validation-derived weighting mainly provides fine-grained adjustment across classifier heads and backbones.

Anatomy-aware temporal event decoding (ATED) appears to be the most influential inference component. Compared with using weighted fusion and per-label event generation without full ATED, adding full ATED substantially improved both mAP@0.5 and mAP@0.95. This supports the design choice of applying temporal smoothing, anatomical mutual exclusivity, monotonic gastrointestinal transit constraints, landmark-region constraints, and morphological refinement before event generation. The comparison between tuple-based and per-label event generation further shows that independent per-label decoding is better aligned with temporal IoU evaluation, because pathological findings should not be split simply because the associated anatomical label changes over time.

The post-competition results further highlight the sensitivity of event-level detection to threshold selection. By extending the original local threshold search with a global coarse search, the post-competition pipeline improved the hidden-test overall score from 0.3530 to 0.3726 at mAP@0.5 and from 0.3235 to 0.3431 at mAP@0.95. Since the learned models and decoding pipeline were unchanged, this improvement suggests that the original F1-initialized local threshold search could miss better event-level operating points for some labels. However, the gain was concentrated on one test video, while the other two videos remained almost unchanged. This indicates that global-augmented threshold search improves robustness in some cases, but does not fully resolve cross-video variation.

Several limitations remain. First, most fusion, threshold, and decoding parameters are still selected on a single validation split, which may limit generalization to unseen videos. Second, DINOv3 remains stronger than EndoFM-LV as a single backbone, suggesting that the current temporal branch may not yet capture long-range video dynamics sufficiently. Third, the post-competition improvement was obtained through inference refinement rather than model retraining, so future work should investigate more robust temporal representation learning, cross-validation-based parameter selection, and decoding strategies that are less sensitive to a single validation split.

\section{Summary}\label{sec5}


Team ACVLab utilized a multi-backbone event detection framework combining EndoFM-LV and DINOv3 ViT-L/16. The combination of BCE with \texttt{pos\_weight}, focal loss, asymmetric loss, class-wise ensemble weighting, and class-specific threshold optimization was utilized to handle class imbalance in the dataset. The results show that structured temporal decoding and per-label event construction are critical for strong event-level performance in capsule endoscopy videos. The team achieved an overall mAP@0.5 of 0.3530, and an overall mAP@0.95 of 0.3235.

After the competition was over, we improved the inference stage of VISTA. The improvement consisted of extending the original local threshold search with a coarse global threshold search while keeping the trained backbones, ensemble heads, fusion strategy, and anatomical decoding pipeline unchanged. This result suggests that event-level performance is sensitive to threshold selection, and that a narrow F1-initialized local search may miss better operating points for some labels. The team achieved an improved overall mAP@0.5 of 0.3726, and an overall mAP@0.95 of 0.3431.

\section{Acknowledgments}\label{sec6}

As participants in the ICPR 2026 RARE-VISION Competition, we fully complied with the competition's rules as outlined in \cite{Lawniczak2025, rarevision2026github}. Our AI model development was based exclusively on the datasets in the competition \cite{LeFloch2025figshareplus}. The mAP values were reported using the test dataset \cite{rarevision_testdata_2026}, and sanity checker \cite{manasapp} released in the competition.

\bibliographystyle{unsrtnat}
\bibliography{sample}

@article{Lawniczak2025,
author = "Anni Lawniczak and Manas Dhir and Maxime Le Floch and Palak Handa and Anastasios Koulaouzidis",
title = "{ICPR 2026 RARE-VISION Competition Document and Flyer}",
year = "2025",
month = "12",
url = "https://figshare.com/articles/preprint/ICPR_2026_RARE-VISION_Competition_Document_and_Flyer/30884858",
doi = "10.6084/m9.figshare.30884858.v3"
}

@misc{manasapp,
  author       = {Manas Dhir and Palak Handa and Anni Lawniczak and Maxime Le Floch},
  title        = {RareEval Socring App},
  year         = {2026},
  howpublished = {\url{https://scoringrarevision.streamlit.app/}},
  note         = {Streamlit application, accessed 2026-03-27}
}

@misc{rarevision2026github,
  author       = "Anni Lawniczak and Manas Dhir and Maxime Le Floch and Palak Handa and Anastasios Koulaouzidis",
  title        = {RARE-VISION-2026-Competition Website},
  year         = {2026},
  howpublished = {\url{https://github.com/RAREChallenge2026/RARE-VISION-2026-Challenge}},
  note         = {Website and GitHub repository for the ICPR 2026 RARE-VISION Competition; accessed 2026-03-27}
}

@article{LeFloch2025figshareplus,
author = "Maxime Le Floch and Fabian Wolf and Lucian McIntyre and Paul Herzog and Christoph Weinert and Albrecht Palm and Konrad Volk and Sophie Helene Kirk and Jonas L. Steinhäuser and Catrein Stopp and Mark Enrik Geissler and Moritz Herzog and Stefan Sulk and Jakob Nikolas Kather and Alexander Meining and Alexander Hann and Jochen Hampe and Nora Herzog and Franz Brinkmann",
title = "{Galar - a large multi-label video capsule endoscopy dataset}",
year = "2025",
month = "2",
url = "https://plus.figshare.com/articles/dataset/Galar_-_a_large_multi-label_video_capsule_endoscopy_dataset/25304616",
doi = "10.25452/figshare.plus.25304616.v2"
}

@misc{rarevision_testdata_2026,
  author       = {Le Floch, Maxime and Lawniczak, Anni and Stopp, Catrein and Zech, Alexander and Kolbig, Alexandra and Tolle, Hannah and Steinhaeuser-Meerz, Jonas L. and Hampe, Jochen and Brinkmann, Franz},
  title        = {Test data for ICPR 2026 - RARE-Vision Competition},
  year         = {2026},
  publisher    = {Technische Universit{\"a}t Dresden},
  doi          = {10.25532/OPARA-1119},
  url          = {https://doi.org/10.25532/OPARA-1119}
}

@article{wang2025improving,
  title={Improving foundation model for endoscopy video analysis via representation learning on long sequences},
  author={Wang, Zhao and Liu, Chang and Zhu, Lingting and Wang, Tongtong and Zhang, Shaoting and Dou, Qi},
  journal={IEEE Journal of Biomedical and Health Informatics},
  year={2025},
  publisher={IEEE}
}

@article{simeoni2025dinov3,
  title={Dinov3},
  author={Sim{\'e}oni, Oriane and Vo, Huy V and Seitzer, Maximilian and Baldassarre, Federico and Oquab, Maxime and Jose, Cijo and Khalidov, Vasil and Szafraniec, Marc and Yi, Seungeun and Ramamonjisoa, Micha{\"e}l and others},
  journal={arXiv preprint arXiv:2508.10104},
  year={2025}
}

@inproceedings{park2023robust,
  title={Robust asymmetric loss for multi-label long-tailed learning},
  author={Park, Wongi and Park, Inhyuk and Kim, Sungeun and Ryu, Jongbin},
  booktitle={Proceedings of the IEEE/CVF international conference on computer vision},
  pages={2711--2720},
  year={2023}
}

@inproceedings{lin2025weighted,
  title={Weighted Stratification in Multi-label Contrastive Learning for Long-Tailed Medical Image Classification},
  author={Lin, Ying-Chih and Chen, Yong-Sheng},
  booktitle={International Conference on Medical Image Computing and Computer-Assisted Intervention},
  pages={677--687},
  year={2025},
  organization={Springer}
}

@article{nam2024deep,
  title={Deep Learning-Based Real-Time Organ Localization and Transit Time Estimation in Wireless Capsule Endoscopy},
  author={Nam, Seung-Joo and Moon, Gwiseong and Park, Jung-Hwan and Kim, Yoon and Lim, Yun Jeong and Choi, Hyun-Soo},
  journal={Biomedicines},
  volume={12},
  number={8},
  pages={1704},
  year={2024},
  publisher={MDPI}
}

@article{lin2024divide,
  title={Divide and conquer: Grounding a bleeding areas in gastrointestinal image with two-stage model},
  author={Lin, Yu-Fan and Qiu, Bo-Cheng and Lee, Chia-Ming and Hsu, Chih-Chung},
  journal={arXiv preprint arXiv:2412.16723},
  year={2024}
}

@inproceedings{lee2025taming,
  title={Taming Domain Shift in Multi-source CT-Scan Classification via Input-Space Standardization},
  author={Lee, Chia-Ming and Qiu, Bo-Cheng and Chen, Ting-Yao and Sun, Ming-Han and Lin, Yu-Fan and Lin, Fang-Ying and Tsai, Jung-Tse and Tsai, I-An and Hsu, Chih-Chung},
  booktitle={Proceedings of the IEEE/CVF International Conference on Computer Vision},
  pages={7331--7339},
  year={2025}
}

\end{document}